\documentclass[10pt,twocolumn,letterpaper]{article}
\pdfoutput=1
\usepackage{wacv}
\usepackage{times}
\usepackage{epsfig}
\usepackage{graphicx}
\usepackage{amsmath}
\usepackage{amssymb}

\usepackage{array}
\usepackage{verbatim}
\usepackage{units}
\usepackage{multirow}
\usepackage{microtype}
\usepackage{booktabs}
\usepackage{amsmath}
\usepackage{amssymb}
\usepackage{graphicx}
\usepackage{bbold}
\usepackage{mathtools}
\usepackage{sistyle}
\usepackage{tabularx}
\usepackage{subcaption}
\SIthousandsep{,}
\usepackage[breakwords]{truncate}

\DeclareRobustCommand{\ctB}[1]{{\color{black}#1}}
\usepackage[accsupp]{axessibility}

\def\wacvPaperID{0143} % Enter the WACV Paper ID here

\wacvfinalcopy % *** Uncomment this line for the final submission

\usepackage[pagebackref=true,breaklinks=true,colorlinks,bookmarks=false]{hyperref}

\begin{document}

%%%%%%%%% TITLE
\title{Addressing out-of-distribution label noise in webly-labelled data}%DSOS: Dynamic Softening of Out-of-distribution Samples for mitigating against web label noise}%
\author{Paul Albert, 
Diego Ortego, Eric Arazo, Noel E. O'Connor, Kevin McGuinness \\
\\
School of Electronic Engineering, \\
Insight SFI Centre for Data Analytics, Dublin City University (DCU) \\
{\tt\small paul.albert@insight-centre.org}
}

\maketitle

\begin{abstract}
   A recurring focus of the deep learning community is towards reducing the labeling effort. Data gathering and annotation using a search engine is a simple alternative to generating a fully human-annotated and human-gathered dataset. Although web crawling is very time efficient, some of the retrieved images are unavoidably noisy, i.e.~incorrectly labeled. Designing robust algorithms for training on noisy data gathered from the web is an important research perspective that would render the building of datasets easier. In this paper we conduct a study to understand the type of label noise to expect when building a dataset using a search engine. We review the current limitations of state-of-the-art methods for dealing with noisy labels for image classification tasks in the case of web noise distribution. We propose a simple solution to bridge the gap with a fully clean dataset using Dynamic Softening of Out-of-distribution Samples (DSOS), which we design on corrupted versions of the CIFAR-100 dataset, and compare against state-of-the-art algorithms on the web noise perturbated MiniImageNet and Stanford datasets and on real label noise datasets: WebVision 1.0 and Clothing1M. Our work is fully reproducible \href{https://git.io/JKGcj}{https://git.io/JKGcj}.
\end{abstract}

%%%%%%%%% BODY TEXT
\section{Introduction}
Deep neural networks (DNNs) are now the standard approach for accurately solving image classification tasks \cite{2019_ICML_EfficientNet,2016_arXiv_Wide}. However, their principal drawback is the large amount of labeled examples required for training. There exist numerous alternatives to deal with the limited availability of labels, such as but not limited to, semi-supervised learning \cite{2021_IJCNN_ReLaB,2020_IJCNN_Pseudo,2019_NeurIPS_MixMatch},~self-supervised learning \cite{2018_ICLR_Rotation,2020_ICML_SimCLR} and robust training on automatically annotated datasets \cite{2017_arXiv_WebVision,2020_ICML_MentorMix}. This paper focuses on the latter.

Designing robust algorithms to train image classification DNNs in the presence of label noise is an important focus for the community \cite{2020_arXiv_LabelNoiseSurvey}; these enable better adaptation of current DNN solutions to real-world problems where extensive curated datasets are unavailable or too expensive to build. Controlled label noise datasets are then often created by synthetically introducing label corruptions in the CIFAR-100~\cite{2009_CIFAR} comparison benchmark. Although good noise robustness is shown on these artificial datasets, web label noise has proven that these solutions generalize poorly to more realistic scenarios and can, in specific cases, be outperformed by robust data augmentation strategies such as mixup~\cite{2020_ICML_MentorMix,2020_ICPR_Robust}. 

We hypothesize that the main limitation for the correction of label noise in web crawled datasets comes from a common assumption made by most label noise robust algorithms \cite {2020_ICLR_DivideMix,2015_ICLR_Bootstrapping,2017_CVPR_ForwardLoss,2020_arXiv_MetaSoftApple} where the true labels for noisy samples lie inside the label set, i.e.~the label noise is \textit{in-distribution} (ID). Conversely, we hypothesize that the label noise present in web crawled datasets is predominantly \textit{out-of-distribution} (OOD), meaning the real labels for noisy samples cannot be inferred from the distribution.
To confirm our hypothesis, we conduct a small but representative survey on the WebVision 1.0 dataset \cite{2017_arXiv_WebVision} to identify the type of noise one can expect in automatically annotated datasets crawled from the web.
\ctB{We then build and validate the DSOS method on controlled corrupted versions of the CIFAR-100 dataset~\cite{2009_CIFAR} where ID noise is introduced using symmetric label flipping and where we use the ImageNet32~\cite{2017_arXiv_IN32} dataset to introduce OOD noise.
We compare with state-of-the-art label noise algorithms on multiple real-world open-source web-crawled datatsets including corrupted versions of the miniImageNet~\cite{2016_NIPS_MiniImageNet} and Stanford Cars~\cite{2013_3dRR13_StanCars} datasets provided by Jiang~\etal~\cite{2020_ICML_MentorMix}, the mini-WebVision dataset~\cite{2017_arXiv_WebVision}, and the Clothing1M~\cite{2015_CVPR_Clothing1M} dataset.
We observe that noisy OOD samples can be leveraged to improve network generalization by enforcing dynamically softening of labels tending to a uniform distribution~\cite{2018_ICLR_confidenceKL} rather than discarding them.

This paper's contributions are: 
\begin{enumerate}
\item We conduct a representative survey over the type of noise to be expected when constructing a dataset using web queries. 
\item We motivate and propose a novel noise detection metric, entropy of the interpolation of the network prediction and the ground-truth label, that is capable to accurately differentiate between clean, ID and OOD noise.
\item We propose DSOS, a simple solution to combat ID and OOD noise in web-crawled datasets and conduct controlled experiments and ablation studies on corrupted versions of the CIFAR-100 dataset.
\item We compare DSOS against state-of-the-art, noise-robust algorithms on real-world web-crawled datasets, demonstrating the validity of our findings for real-world applications.
\end{enumerate}}

\section{Related work}

\subsection{Label noise detection}
Label noise detection aims at distinguishing between clean and noisy samples in an unsupervised manner. The commonly used method is the small loss trick \cite{2019_ICML_BynamicBootstrapping,2019_ICML_SELFIE,2018_CVPR_JointOpt}, which is based on the assumption that when training a neural network with a high learning rate, noisy samples will have a higher loss than their clean counterpart. The small loss observation extends to other metrics such as forgetting event count \cite{2019_ICLR_Forgetting}, pre-trained mentor network scoring \cite{2018_ICML_MentorNet}, uncertainty \cite{2019_CVPRW_Uncertainty}, prediction consistency \cite{2019_ICML_SELFIE}, accuracy, or entropy. The small loss can also be applied in multiple network settings to improve the detection \cite{2020_ICLR_DivideMix,2018_NeurIPS_CoTeaching}. Other noise detection algorithms include using the Local Outlier Factor \cite{2018_CVPR_IterativeNoise} to identify isolated samples in the feature space or meta-learning \cite{2020_arXiv_MetaSoftApple}. \ctB{Training neural networks to detect OOD examples could also be considered relevant to creating a label noise robust algorithm, but this approach systematically requires at least a trusted, exclusively ID dataset \cite{2019_NeurIPS_likelihoodood,2019_ICCV_oodpseudounsup} and sometimes an additional exclusively OOD dataset \cite{2018_ECCV_oodselfsup}. This constraint is too limiting in scenarios where the nature of the noise is unknown.}

\subsection{Algorithms robust to label noise}
DNNs have been shown to easily overfit noisy labels, leading to generalization degradation \cite{2017_ICLR_Rethinking}. We categorize the first class of label noise robust algorithms as label correction algorithms. The goal for label correction algorithms is to denoise the dataset by guessing the true label for noisy samples. We include here approaches that perform this correction online when computing the final loss. Label guessing strategies include: label transition matrices \cite{2017_CVPR_ForwardLoss}, current network predictions \cite{2019_ICML_BynamicBootstrapping,2015_ICLR_Bootstrapping}, semi-supervised learning \cite{2018_WACV_SemiSupNoise,2020_ICPR_Robust}, and meta-learning inspired backpropagation \cite{2020_arXiv_MetaSoftApple}. The second correction strategy is centered around limiting the contributions of the noisy labels to the network's parameters by either using a curriculum \cite{2018_ICML_MentorNet,2018_ECCV_CurrNet} or contribution weights in the final loss \cite{2019_NeurIPS_MetaWeight} that diminish the contribution of noisy samples in the gradient update. Other strategies include pushing apart the representations of clean and noisy samples in the feature space \cite{2018_CVPR_IterativeNoise} or noise robust data augmentation \cite{2018_ICLR_mixup}. \ctB{Two algorithms have recently been proposed to tackle separate ID and OOD retrieval. EvidentialMix~\cite{2020_WACV_EDM} proposes a separate detection of ID and OOD samples using the evidiential loss~\cite{2018_NeurIPS_evidentialloss} but chooses to ignore the OOD samples to increase accuracy on the ID noise correction using the DivideMix~\cite{2020_ICLR_DivideMix} algorithm. JoSRC~\cite{2021_CVPR_JoSRC} proposes to differentiate between OOD and ID samples using a contrastive evaluation using multiple views of the same noisy sample. JoSRC additionally proposes a fixed smoothing for the labels of detected OOD samples using a fixed temperature hyperparameter. Both of these algorithms additionally require two networks.}
Recent studies \cite{2020_ICML_MentorMix,2020_ICPR_Robust} show that the improvements noise robust algorithm observe on synthetic datasets do not always translate to realistic label noise scenarios.

\begin{table}[t]
\caption{\label{tab:noiseinweb} Analysis on the noise types and ratios found in mini-WebVision. We randomly sample three subsets (S) of 2000 images and report correctly-labeled samples and in-distribution (ID) and out-of-distribution (OOD) noisy samples. Image examples are available in the supplementary material.}
\centering{}
\global\long\def\arraystretch{0.9}%
\resizebox{0.8\columnwidth}{!}{{{}}%
\begin{tabular}{l>{\centering}c>{\centering}c>{\centering}c>{\centering}c}
\toprule
 & {S1} & {S2} & {S3} & {Average (\%)}\tabularnewline
\midrule
{Correct} & $1441$  & $1440$ & $1335$ & $1405.33~(70.30)$\tabularnewline
{OOD} & $460$ & $429$ & $573$ & $487.33~(24.38)$\tabularnewline
{ID} & $98$ & $130$ & $91$ & $106.33~(5.32)$ \tabularnewline
\bottomrule
\end{tabular}}
\end{table}

\begin{figure*}
\centering
\includegraphics[width=.9\linewidth]{"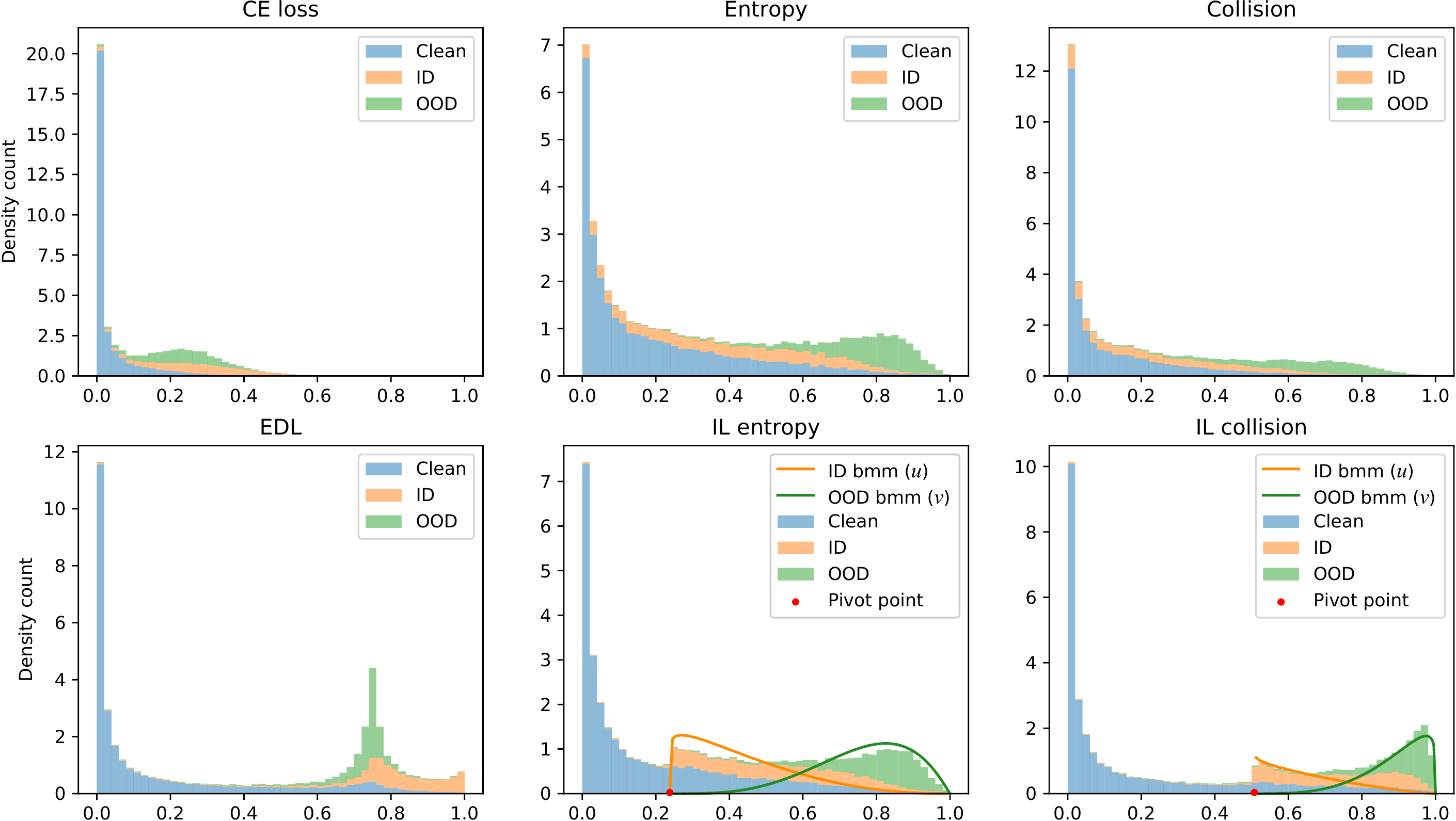"}
\par
\caption{Stacked density histograms for multiple noisy sample retrieval measures on CIFAR-100 with $\rho=\psi=0.2$. All metrics are min-max normalized. For the entropy of the intermediate label (IL) we also draw the decision function (BMM) that we fit to the data. The pivot point in red separates clean from noisy samples. \label{fig:metricc10}}
\end{figure*}
\section{Web datasets and out-of-distribution noise\label{sec:expstudy}}
Recent state-of-the-art for label noise detection and correction rely on strong assumptions verified on synthetically generated noise. Recent contributions~\cite{2020_ICML_MentorMix,2020_ICPR_Robust} demonstrated that many algorithms developed on synthetic datasets do not generalize well to real-world label noise and that improvements are often inferior to using data augmentation (mixup~\cite{2018_ICLR_mixup}). We suggest that this limitation is a consequence of a strong assumption made by noise-robust algorithms where noisy samples have their labels corrected by assigning another label from the known label distribution, i.e.~the noise is in-distribution. We conversely hypothesize that most of the noise in web labeled datasets is out-of-distribution, meaning the real unknown label lies outside of the known label set. To verify this hypothesis we randomly sample images from the real-world label noise dataset mini-WebVision (first 50 classes subset of the WebVision 1.0 dataset \cite{2017_arXiv_WebVision}) and manually categorize their label in three categories: clean, in-distribution noise, and out-of-distribution noise. We separate the labeling process in two steps: a first round on the images to detect any image whose label does not correspond to the class to which it is assigned, and a second pass on the noisy images alone to classify them as ID when the true label lies in the known set of classes, or OOD when it does not. We repeat the process for three random subsets of the mini-WebVision dataset. Table \ref{tab:noiseinweb} shows the results of the study, demonstrating the clear domination of out-of-distribution noise over in-distribution noise \ctB{(A visualization of the noise categorization of the images is available in the supplementary material).} This observation sheds light on the limited improvements of in-distribution label correction techniques when applied to web crawled datasets, while explaining the benefits of undersampling algorithms, which sample noisy data less often to reduce their contribution~\cite{2018_ECCV_CurrNet,2020_ICML_MentorMix,2020_WACV_EDM} at the cost of ignoring a part of the data.

\section{DSOS}
Taking into consideration the results observed in Section \ref{sec:expstudy}, we propose Dynamic Softening of Out-of-distribution Samples (DSOS), a label correction algorithm for robust learning on web label noise distributions.
We aim to solve an image classification task over $C$ classes as learning a DNN model $h_{\psi}$ given a training set $\mathcal{D} = \left\{ \left(x_{i},y_{i}\right)\right\} _{i=1}^{N}$
of $N$ samples where $x_i \in \mathcal{X}$. More specifically, we tackle the case where the dataset consists of a correctly labeled set $\mathcal{D}_{c}=\left\{ \left(x_{i},y_{i}\right)\right\} _{i=1}^{N_{c}}$
with corresponding one-hot encoded labels $y_{i}\in\left\{ 0,1\right\} ^{C}$, an incorrectly labeled in-distribution noisy set $\mathcal{D}_{in}=\left\{ \left(x_{i},y_{i}\right)\right\}_{i=1}^{N_{in}}$ and of an out-of-distribution noisy set $\mathcal{D}_{out}=\left\{ \left(x_{i},y_{i}\right)\right\}_{i=1}^{N_{out}}$.
We denote $N=N_{c}+N_{in}+N_{out}$ the total number of available samples. We consider unknown the distribution of the samples between $\mathcal{D}_{c}$, $\mathcal{D}_{in}$ and $\mathcal{D}_{out}$.
We note $h:\mathcal{X\rightarrow}\left[0,1\right]^{C}$ the deep neural network (DNN) we train to classify the images as belonging to a class $c \in \{1,\ldots, C\}$.

\ctB{\subsection{Separate detection of ID and OOD noise\label{sec:metchoice}}
\subsubsection{Motivation}
We motivate here the need for a new metric for the dual detection of ID and OOD noise in web crawled datasets by considering the ideal case where a network has been trained on a web-crawled dataset and did not overfit  the noise. Samples would then be characterized by either a confident correct prediction (clean samples), a confident incorrect prediction (ID noise), or an un-confident prediction (OOD noise). Using a DNN to detect noisy samples, metrics from in the label noise literature propose to either quantify the accuracy of the prediction~\cite{2019_ICML_BynamicBootstrapping, 2020_ICLR_DivideMix, 2018_NeurIPS_CoTeaching} (cross-entropy loss, accuracy, Kullback-Leibler divergence) or the uncertainty of the prediction~\cite{2019_ICLR_Forgetting,2021_CVPR_JoSRC} (forgetting events, entropy of the prediction, contrastive predictions). Relying on one characterization of the network prediction alone is problematic when presented with the duality of the noise present in web-crawled datasets as ID and OOD noise cannot be independently retrieved. While accuracy approaches indistinguishably retrieve incorrectly predicted OOD and ID noise (both having low agreement with their noisy label), certainty-based approaches only retrieve under-confident OOD noise. EvidentialMix~\cite{2020_WACV_EDM} proposes an independent retreival of ID and OOD noise, where a mean square error + variance loss~\cite{2018_NeurIPS_evidentialloss} (evidential loss, EDL) is shown to separate ID and OOD noise on artificial corrupted noisy datasets (CIFAR-10~\cite{2009_CIFAR}. We argue that the limitation of the evidential loss for web-crawled datasets lies in the absence of separation between OOD noise and lower-confidence predictions in general, resulting in a sub-optimal OOD retrieval, the dominant noise type for web-crawled datasets. This limitation is evidenced in Figure~\ref{fig:metricc10} (described in Section~\ref{sec:newmetric}) and in Table~\ref{tab:retcomp} where we compare retrieval scores for Clean/ID/OOD samples (one versus all) for an accuracy (CE loss) or confidence metric (entropy) against using the EDL loss fitted with a 3 components Gaussian mixture model~\cite{2020_WACV_EDM}, and two variations of our proposed metric (see Section~\ref{sec:newmetric}). The table highlights the trade-off we make for  better OOD detection at the cost of less accurate ID retrieval when compared with EDL.

\begin{table}[t]
\caption{\label{tab:retcomp} AUC retreival score for different types of metrics after warm-up on CIFAR-100 with $\rho = \psi = 0.2$}
\centering{}
\global\long\def\arraystretch{0.9}%
\resizebox{0.7\columnwidth}{!}{{{}}%
\begin{tabular}{l>{\centering}c>{\centering}c>{\centering}c>{\centering}c}
\toprule
 & {Clean} & {ID} & {OOD}\tabularnewline
\midrule
{Small loss} & $95$  & $87$ & $81$\tabularnewline
{EDL} & $93$ & $90$ & $75$ \tabularnewline
{IL entropy} & $91$ & $81$ & $94$ \tabularnewline
{IL collision} & $93$ & $85$ & $92$ \tabularnewline
\bottomrule
\end{tabular}}
\end{table}

\begin{figure*}
\centering
\includegraphics[width=.9\linewidth]{"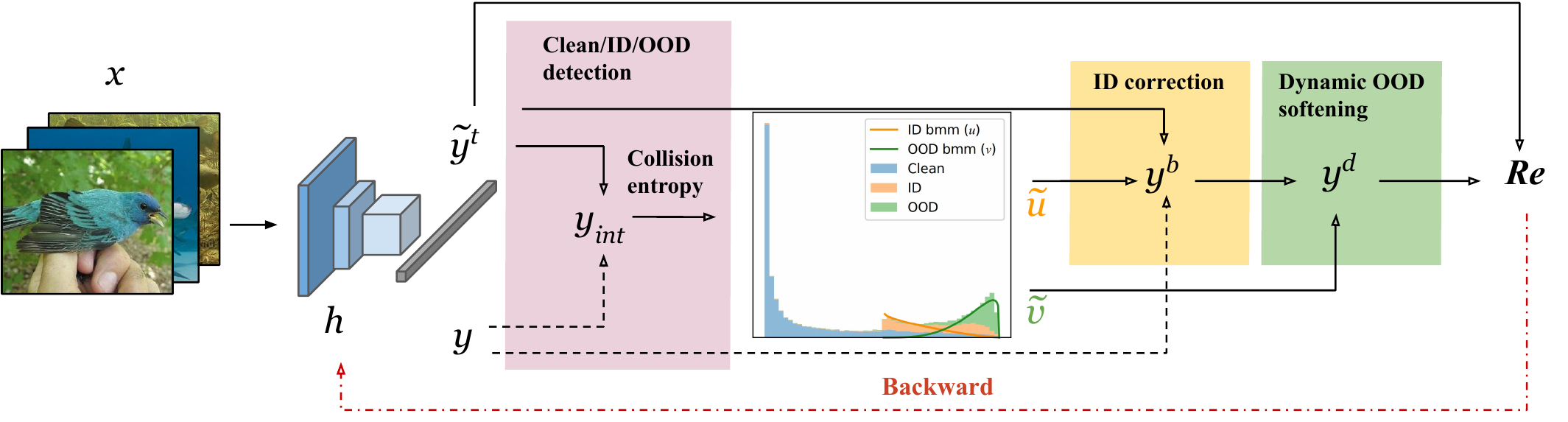"}
\par
\caption{Visualization of the DSOS algorithm. DSOS identifies and corrects the ID and OOD noise from the training distribution before applying targeted label correction. \label{fig:dsos}}
\end{figure*}

\subsubsection{Metric~\label{sec:newmetric}}
We propose a novel noise detection metric that allows the separate detection of confident clean samples, confident ID noisy samples, and OOD noisy samples. To do so, we propose to compute the intermediate label between the current network prediction
$\tilde{Y}$ and the target label $Y$: $y_{int} = \frac{y_i+\tilde{y_i}}{2}$ and to study its collision entropy:
\begin{align}
    \mathit{l}_\mathit{detect} = -\log{\left(\sum_{c=1}^{C}{y_{\mathit{int},c}^2}\right)}.
\end{align}
We aim to detect three different events for $y_\mathit{int}$: the clean event where prediction and ground truth agree, resulting in a low entropy; the ID event where prediction and ground truth are both confident but disagree (medium entropy); and the OOD event where the prediction is under-confident (high entropy). Studying the entropy of the intermediate label $\mathit{l}_\mathit{detect}$ allows us to reverse the detection hierarchy observed in the EDL from clean-OOD-ID to clean-ID-OOD since confident incorrect predictions are now observed in $y_\mathit{int}$ as a bimodal distribution that has a lower entropy than an interpolation of the ground truth with an un-confident uniform prediction. A fundamental property of $\mathit{l}_\mathit{detect}$ is that it differentiates between low confidence but correct predictions (clean samples) and confident incorrect predictions (ID noise), which is evidenced by the pivot point. The pivot point is defined for $y_{int}$ being a perfect bi-model distribution, i.e.~two high probability modes with values $0.5$ with all other bins to $0$, resulting in $\mathit{l}_\mathit{detect} = -\log{0.5}$, the pivot point. Detecting these events of high probability motivate our choice of using the collision entropy, which is more sensitive to high probability events than the Shannon entropy.
Using the pivot point together with the observed bimodality of the noisy samples, we classify the samples in three distinct categories where every sample whose $\mathit{l}_\mathit{detect}$ value is inferior to the pivot point is considered clean and where we fit a two components Beta Mixture Model (BMM) to the noisy samples. By computing the posterior probability of a sample to belong to each component, we evaluate the ID and OOD nature of every noisy sample.

Figure~\ref{fig:metricc10} illustrates the clean/ID/OOD separation  observed for accuracy based and uncertainty based metrics on the CIFAR-100 dataset corrupted with 20\% symetric ID noise and 20\% OOD noise from ImageNet32~\cite{2017_arXiv_IN32} at the end of the warm-up phase (see Section~\ref{sec:expsetup} for training details). The figure illustrates how the collision entropy improves the separation between clean and ID noise over the Shannon entropy and how we trade off improved OOD detection for a decreased ID detection over the evidential loss (EDL)~\cite{2020_WACV_EDM} (see Table~\ref{tab:retcomp}). The pivot point is indicated in red. An additional illustration explaining the behavior of $\mathit{l}_\mathit{detect}$ for intermediate configurations of $y_\mathit{int}$ is available in the supplementary material.}

\subsection{DSOS~\label{sub:DSOStheo}}
We build DSOS as a single network based, single training cycle algorithm which aims to first discover ID and OOD samples in a corrupted dataset before separately addressing ID and OOD noise using dynamic label correction strategies. Figure~\ref{fig:dsos} illustrates the DSOS algorithm. We aim to correct ID samples using confident predicted label assignments and to promote high entropy prediction for OOD samples which cannot be corrected.
DSOS aims to minimize the following empirical risk over the noisy dataset:
\begin{align}
\mathit{R_e} = \frac{1}{N} \sum_{i=1}^{N} -{y_i^t}^T\log{h(x_i)},
\end{align}
where the logarithm is applied element-wise and $y_i^t$ denotes the, possibly unknown, true label for sample $x_i$.
Although it is possible to directly minimize $\mathit{R_e}$ for ID noisy samples by correcting the noisy label $y_i$ to the true label $y_i^t$, this is not the case for OOD label noise. We propose then not to attempt to approximate the true label of OOD samples using a label from the known distribution but instead to promote better network calibration by encouraging high-entropy predictions, i.e.~a uniform prediction over ID classes. 
We then rewrite empirical risk as:
\begin{align}
\begin{split}
   \mathit{R_e} = & -\frac{1}{N_c+N_{in}} \sum_{i=1}^{N_c+N_{in}} {y_i^t}^T\log{h(x_i)} \\
   & - \frac{1}{N_{out}} \sum_{j=1}^{N_{out}} {y_s}^T\log{h(x_j)}, 
\end{split}
\end{align}
where $y_s$ is the softened label, i.e.~a perfect uniform prediction over all the classes C.
To obtain a dynamic softening from $y_i^t$ to $y_s$ and given a OOD classifier $\mathcal{V} = \{v_{i}\}_{i=1}^{N}, v_i \in [0,1]$ where $v_i=0$ means sample $x_i$ is OOD, we minimize:
\begin{align}
\begin{split}
   \mathit{R_e} = & -\frac{1}{N} \sum_{i=1}^{N} f({y_i^t}, v_i)^T\log{h(x_i)},
   \label{eq:risk}
\end{split}
\end{align}
with $f({y_i^t}, v_i)$ the smoothing function where $f({y_i^t}, 0) = y_s$ and $f({y_i^t}, 1) = {y_i^t}$.
\subsubsection{Label softening of out-of-distribution samples}
We minimize the risk in Eq.~\ref{eq:risk} using a label correction approach where we aim to first correct the labels for noisy ID samples to their true label using a bootstrapping inspired approach \cite{2019_ICML_BynamicBootstrapping,2015_ICLR_Bootstrapping,2019_ICML_SELFIE}. For the OOD samples, we propose a dynamic softening strategy by computing the cross-entropy loss with regards to a dynamically smoothed label (the more likely a sample is detected to be OOD, the more uniform the target) and avoid using an additional regularization term (Kullback-Leibler divergence minimization between the prediction and a uniform target would be a common solution~\cite{2018_ICLR_confidenceKL}).
To correct ID label noise, we consider a first estimated metric $\tilde U = \{\tilde{u}_i\}_{i=0}^N$, where $\tilde{u}_i \in \{0,1\}$, evaluating whether a sample is noisy but in-distribution, i.e.~the label can be corrected to another from the distribution. $\tilde{u}_i = 1$ denotes sample $x_i$ is noisy but ID. We denote $\tilde{y}_i^t$ the current true label guess for sample $x_i$ an correct it with,
\begin{align}
    y_i^b = (1 - \tilde{u}_i) y_i + \tilde{u}_i \tilde{y}_i^t. 
\end{align}
Regarding OOD label noise, we consider a second metric $\tilde V = \{\tilde{v}_i\}_{i=0}^N$ estimating $\mathcal{V}$ and evaluating whether a sample is noisy and OOD ($\tilde{v}_i \in \left(0,1\right]$) with $v_i=0$ meaning a sample is considered OOD. We re-normalize the possibly bootstrapped label $y_i^b$ for a sample $x_i$ assigned to an OOD noisiness metric estimation $\tilde{v}_i$ as 
\begin{align}
y_{i}^{d} = \frac{\exp{\frac{\tilde{v}_iy_i^b}{\alpha }}}{\sum_{c=1}^C \exp{\frac{\tilde{v}_iy_{i,c}^b}{\alpha }}}.
\end{align}
with $\alpha \in [0,1]$ a hyperparameter.
$y_{i}^{d}$ is a dynamically smoothed correction of the corrected label $y_i^b$ where $\frac{\tilde{v}_i}{\alpha}$ serves as a dynamic temperature depending on the out-of-distribution noisiness of the sample. In Figure~\ref{fig:metricc10}, $\tilde{U}$ corresponds to the posterior probability given $\mathit{l}_\mathit{detect}$ for the left-most beta mixture being superior to $0.5$ and $\tilde{V}$ is the posterior probability of the right-most beta mixture given $\mathit{l}_\mathit{detect}$ (no threshold). We evaluate $\tilde{U}$ and $\tilde{V}$ every epoch starting at the end of the warm-up phase where the network is trained without correction on the noisy dataset. We end the warm-up phase one epoch after the first learning rate reduction.
In summary, OOD noisy labels will be dynamically replaced by a uniform distribution hence promoting their rejection by the network and the clean and corrected ID noisy samples will be assigned a moderately smoothed label, which has been proven to be beneficial for robust DNN training in the presence of label noise \cite{2020_ICMLW_SmoothLabelOOD,2020_ICML_SmoothLabelnoise}. Both $\tilde{U}$ and $\tilde{V}$ are cut of from the computation graph and neither is backpropageted in equation~\ref{eq:risk}.

\begin{table*}[t]
    \caption{DSOS for mitigating ID and OOD noise on CIFAR-100 corrupted with ImageNet32 images. We run each algorithm with the exact same noise corruption. We report best and last accuracy (best/last).
    \label{tab:sotac100}}
    \global\long\def\arraystretch{0.9}%
    \centering
    \resizebox{1\textwidth}{!}{%
    \centering
    \begin{tabular}{c>{\centering}c>{\centering}c>{\centering}c>{\centering}c>{\centering}c>{\centering}c>{\centering}c>{\centering}c>{\centering}c>{\centering}c}
    \toprule
    $\rho$ & $\psi$ & CE & M & DB & ELR & EDM & JoSRC & \multicolumn{3}{c}{DSOS}  \tabularnewline
    \cmidrule(lr){9-11}
     & & & & & & & & ID & OOD & both \tabularnewline
    \midrule
    0.2 & 0.2 & 63.68/55.52 & 66.71/62.52 & 65.61/65.61 & 63.90/63.72 & 65.11/64.49 & 67.37/64.17 & 68.09/67.78 & 69.37/69.37 & 70.54/70.54 \tabularnewline
    0.4 & 0.2 & 58.94/44.31 & 59.54/53.16 & 54.79/54.42 & 57.16/56.91 & 55.65/54.49 & 61.70/61.37 & 60.12/59.32 & 62.34/61.03 & 62.49/62.05 \tabularnewline
    0.6 & 0.2 & 46.02/26.03 & 42.87/40.39 & 42.50/42.50 & 31.20/29.55 & 28.51/10.47 & 37.95/37.11 & 46.10/42.93 & 46.54/40.23 & 49.98/49.14 \tabularnewline
    0.4 & 0.4 & 41.39/18.45 & 38.37/33.85 & 35.90/35.90 & 22.85/21.63 & 24.15/01.62 & 41.53/41.44 & 40.94/35.89 & 42.53/39.76 & 43.69/42.88 \tabularnewline
       \bottomrule
    \end{tabular}}
\end{table*}  

\ctB{\subsubsection{Additional regularization}
In order to be competitive with the state-of-the-art, we pair DSOS with two different regularization strategies commonly used to combat label noise. The first regularization we add to the loss promotes high-entropy predictions on ID samples:
\begin{align}
    \mathit{l}_{e} = - \frac{1}{N}\sum_{i=1}^N\tilde{v_i}\sum_{i=1}^N{h(x_i)\log(h(x_i))}.
\end{align}
We find $\mathit{l}_{e}$ to be especially important in the warm-up phase as it promotes confident predictions for both the clean samples and the ID samples, which enables better detection. During the label correction phase of DSOS, the regularization is proportionally weighted according to the clean and noisy ID samples detection $\tilde{V}$ so as to not to go against the label softening strategy for OOD samples.
We additionally pair DSOS with mixup~\cite{2018_ICLR_mixup} data augmentation, which has shown to be robust to label noise and that is commonly used in related state-of-the-art noise robust approaches. An ablation study for the different components of DSOS including the effect of the regularizations is given in Section~\ref{sec:abla}.
The final loss DSOS minimizes is:
\begin{align}
\mathit{l} = -\frac{1}{N} \sum_{i=1}^N {y^d}^T\log(h(x_i)) + \gamma \mathit{l}_e
\end{align}
with $\gamma=0.4$.
}

\begin{table}[]
    \caption{Ablation study for DSOS. We report best and last accuracy.\label{tab:abla}}
    \global\long\def\arraystretch{1}%
    \centering
    \resizebox{.4\textwidth}{!}{{{}}%
    \begin{tabular}{l>{\centering}c>{\centering}c}
    \toprule
        & Best & Last\tabularnewline
        \midrule
       CE  & 63.68 & 55.52 \tabularnewline
       + mixup & 66.71 & 62.52 \tabularnewline
       + Entropy regularization & 67.27 & 63.04 \tabularnewline
       + Batch normalization tuning & 67.56 & 65.69 \tabularnewline
       + In-distribution bootstrapping & 68.09 & 67.78  \tabularnewline
       + Out-of-distribution softening & 70.54 & 70.54 \tabularnewline
       \bottomrule
    \end{tabular}}
\end{table}

\begin{table*}[t]
    \caption{Comparison of DSOS with state-of-the-art algorithms on MiniImageNet and Stanford Cars corrupted with web label noise gathered by \protect\cite{2020_ICML_MentorMix} (red noise). We bold best and underline last accuracy for the best performing algorithm. 
    \label{tab:sotastan}}
    \global\long\def\arraystretch{0.9}%
    \resizebox{1\textwidth}{!}{%
    \centering
    \begin{tabular}{l>{\centering}c>{\centering}c>{\centering}c>{\centering}c>{\centering}c>{\centering}c>{\centering}c>{\centering}c>{\centering}c}
    \toprule
    Dataset & Noise level & CE & D & SM & B & M & MN & MM & DSOS\tabularnewline
    \midrule
    MiniImageNet & $0$ & $70.9/68.5$ & $71.8/65.7$ & $71.4/68.4$ & $71.8/68.4$ & $72.8/72.3$ & $71.2/68.9$ & $74.3/73.7$ & $\textbf{74.52}/\underline{74.10}$ \tabularnewline
        & $30$ & $66.1/56.5$ & $66.6/55.0$ & $65.2/56.3$ & $66.6/56.7   $ &$66.8/61.8$ & $66.2/64.0$ & $68.3/67.2$ & $\textbf{69.84}/\underline{67.86}$\tabularnewline
        & $50$ & $60.9/51.7$ & $62.1/50.01$ & $61.3/51.3$ & $62.6/52..5$ & $63.2/58.4$ & $61.7/58.0$ & $63.3/61.8$ & $\textbf{66.14}/\underline{65.18}$\tabularnewline
        & $80$ & $48.8/39.8$ & $49.5/37.6$ & $49.0/40.6$ & $50.1/40.1$ & $50.7/45.5$ & $49.3/43.4$ & $50.2/48.4$ & $\textbf{55.26}/\underline{52.24}$ \tabularnewline
       \midrule
    Stanford Cars & $0$ & $90.8/90.8$ & $\textbf{92.2}/\underline{92.2}$ & $90.1/90.1$ & $90.3/90.0$ & $91.9/91.9$ & $90.2/90.1$ & $91.8/91.6$ & $91.38/91.27$ \tabularnewline
       & $30$ & $80.4/80.2$ & $87.6/87.6$ & $82.2/81.9$ & $83.4/83.0$ & $85.6/85.2$ & $81.1/80.9$ & $87.8/87.7$ & $\textbf{88.36}/\underline{88.14}$ \tabularnewline
       & $50$ & $70.6/70.3$ & $79.3/79.2$ & $70.1/70.1$ & $73.6/73.5$ & $79.1/78.9$ & $72.0/72.0$ & $80.4/79.8$ & $\textbf{82.04}/\underline{81.72}$ \tabularnewline
       & $80$ & $43.3/43.0$ & $61.8/61.8$ & $46.4/46.4$ & $47.4/46.7$ & $55.7/55.4$ & $51.0/50.9$ & $58.6/58.6$ & $\textbf{62.36}/\underline{62.36} $ \tabularnewline
       \bottomrule
    \end{tabular}}
\end{table*}

\section{Experiments}
\subsection{Experimental setup~\label{sec:expsetup}}
\ctB{We conduct controlled experiments on corrupted versions of the CIFAR-100 dataset~\cite{2009_CIFAR} using ImageNet32~\cite{2017_arXiv_IN32} images for the OOD noise. The CIFAR-100 dataset is a $32\times32$ image dataset composed of $50.000$ training images and $10.000$ test images, equally distributed over $100$ classes. The ImageNet32 dataset is a $32\times32$ downsized version of the ILSVRC12~\cite{2012_NeurIPS_ImageNet} dataset ($1.000$ classes and $1.2$M images). In order to corrupt CIFAR-100, we consider the OOD noise ratio $\rho$ and the ID noise ratio $\psi$. We first replace a random fraction $\rho$ of the CIFAR-100 images by randomly selected ImageNet32~\cite{2017_arXiv_IN32} images and randomly flip a $\psi$ fraction of the clean samples to a random label assignment. The total noise ratio is $\psi + \rho$. We train for $100$ epochs, using a PreActivation ResNet18~\cite{2016_CVPR_ResNet}, SGD with momentum $0.9$ and weight decay $5\times10^{-4}$, starting from a learning rate of $0.03$ and reducing it by $10$ at epochs $50$ and $80$, batch size $32$ ($64$ for the warm-up).

For controlled web-crawled datasets, we consider different noise levels ($0\%, 30\%, 50\%, 80\%$) for the web label noise corruption released for the MiniImageNet ($50k$ training images, $10.000$ test images) and StanfordCars ($8k$ training images, $8k$ test images) datasets~\cite{2020_ICML_MentorMix}, adopting the 299$\times$299 image resolution for training and the InceptionResNetV2 network architecture. We train for $200$ epochs, using SGD with momentum $0.9$ and weight decay $5\times10^{-4}$, starting from a learning rate of $0.01$ and reducing it by $10$ at epochs $100$ and $160$, batch size $32$.
For real-world web-crawled datasets, we report results training on the mini-Webvision~\cite{2017_arXiv_WebVision} dataset (first 50 classes of WebVision) ($66k$ training images, $2.5k$ test images) at resolution $224\times224$. We train for $100$ epochs, using an InceptionResNetV2, SGD with momentum $0.9$ and weight decay $5\times10^{-4}$, starting from a learning rate of $0.01$ and reducing it by $10$ at epochs $50$ and $80$, batch size $32$. We use the mini-WebVision validation set for early stopping and the ILSVRC12 dataset~\cite{2012_NeurIPS_ImageNet} as a test set. For Clothing1M~\cite{2015_CVPR_Clothing1M} ($1M$ training images, $15k$ test images) we sample 1000 random batches every epoch, resolution $227\times227$. We train for $100$ epochs using a ResNet50 pretrained on ImageNet, SGD with momentum $0.9$ and weight decay $1\times10^{-3}$, starting from a learning rate of $0.002$ and reducing it by $10$ at epochs $50$ and $80$, batch size $32$. The dataset configurations and networks used follows the state-of-the-art we compare with~\cite{2020_ICML_MentorMix,2020_ICLR_DivideMix,2020_NeurIPS_EarlyReg}. 
A summary of the training details is available in the supplementary material.}

\begin{table*}[t]
    \centering
    \caption{Classification accuracy for DSOS and state-of-the-art methods against methods using a unique network vs an ensemble. We train the network on the mini-Webvision dataset and test on the Imagenet 1k test set (ILSVRC12). All results except our own (DSOS) are from \protect\cite{2020_NeurIPS_EarlyReg}. We bold the best results. \label{tab:sota}}
    \global\long\def\arraystretch{0.9}%
    \resizebox{.85\textwidth}{!}{{{}}%
    \begin{tabular}{l>{\centering}c>{\centering}c>{\centering}c>{\centering}c>{\centering}c>{\centering}c>{\centering}c>{\centering}c>{\centering}c>{\centering}c}
    \toprule
    	& & \multicolumn{6}{c}{Unique network} & \multicolumn{3}{c}{Ensemble of two networks} \tabularnewline
	\cmidrule(lr){3-8}\cmidrule(lr){9-11}
        & & F & Co-T & M & MM & ELR & DSOS & DM & ELR+ & DSOS \tabularnewline
    \midrule
       \multirow{2}{*}{mini-WebVision} & top-1 & $61.12$ & $63.58$ & $75.44$ & $76.0$ & $76.26$ & \textbf{77.76} & $77.32$ & $77.78$ & \textbf{78.76}\tabularnewline
        & top-5 & $82.68$ & $85.20$ & $90.12$ & $90.2$ & $91.26$ & \textbf{92.04} & $91.64$ & $91.68$ & \textbf{92.32}\tabularnewline
       \multirow{2}{*}{ILSVRC12} & top-1 & $57.36$ & $61.48$ & $71.44$ & $72.9$ & $68.71$ & \textbf{74.36} & $75.20$ & $70.29$ & \textbf{75.88} \tabularnewline
       & top-5 & $82.36$ & $84.70$ & $89.40$ & \textbf{91.10} & $87.84$ & $90.80$ & $90.84$ & $89.76$ & \textbf{92.36}\tabularnewline
       \bottomrule
    \end{tabular}}
\end{table*}

\ctB{\subsection{Experiments on CIFAR-100~\label{sec:controled}}
We test DSOS in a controlled noise scenario on the CIFAR-100 dataset corrupted with ID symetric label noise and OOD images from the ImageNet32 dataset in Table~\ref{tab:sotac100}. Contrary to previous works~\cite{2021_CVPR_JoSRC}, the focus here is on OOD noise. We consider 4 different configurations for CIFAR-100 with $\rho \in [0.2, 0.4, 06]$ and $\psi \in [0.2, 0.4]$. We show the benefits of DSOS when performing ID label bootstrapping or OOD label softening alone as well as the combined benefits of the dual label correction (both in Table~\ref{tab:sotac100}). We compare our approach with two simple baselines: CE, a simple cross-entropy training without any noise correction and mixup (M)~\cite{2018_ICLR_mixup} a data augmentation strategy robust to label noise. We additionally report results for state-of-the-art noise robust algorithms including Dynamic Bootstrapping (DB)~\cite{2019_ICML_BynamicBootstrapping} and Early Learning Regularization (ELR)~\cite{2020_NeurIPS_EarlyReg}. Finally, we run algorithms focused on OOD and ID noise robustness: EvidentialMix (EDM)~\cite{2020_WACV_EDM} and JoSRC~\cite{2021_CVPR_JoSRC}.
We use the same hyperparameters and network as ours for training the algorithms we compare with except for JoSRC which uses the Adam optimizer by default.
For DSOS, we perform a warm-up training up until after the learning rate reduction. One epoch after the learning rate reduction, we start performing ID and OOD noise detection and apply our label correction strategy with $\alpha=0.05$. We find that performing warm-up with mixup (M) is better as long as the total noise is superior to $0.8$ but use a simple CE warm-up for total noise levels of $0.8$. We systematically use the entropy regularization term for the warm-up phase. We report running DSOS with ID or OOD correction alone as well as with both correction (both). If we notice that the BMM does not capture the ID mode (mode of the first beta distribution outside of the $[0, 1]$ interval) which we observe for total noise levels of $0.8$, we fall back to using $\mathit{l}_\mathit{detect}$ directly for detecting the ID noisy samples ($\mathit{l}_\mathit{detect} < 0.5$ means a samples is ID noisy). We draw the attention of the reader to the improvements DSOS brings when compared to other ID/OOD noise correction approaches even though we use a single network.}

\subsection{Ablation study~\label{sec:abla}}
We conduct an ablation study to highlight the important elements of DSOS trained on CIFAR-100 with $\rho=0.2$ and $\psi=0.2$ (Table~\ref{tab:abla}). 
We find entropy regularization \cite{2018_CVPR_JointOpt} to be necessary to promote confident predictions and specifically study the case where the metrics tracking and the bootstrapped label predictions necessary to applying ID noise correction are computed with trainable batch normalization layers, i.e.~the layers get tuned with unmixed samples before evaluation on the validation set.
The ablation study highlights how the introduction of the dynamic label softening strategy improves accuracy results over applying ID label correction alone.

\begin{table*}[t]
    \centering
    \caption{Comparison of DSOS against state-of-the-art algorithms on Clothing1M. Top-1 best accuracy on the test set. We run ELR+ and DM using the code provided by the authors. All other results are from the specified works. We bold the best results. \label{tab:sotaclothing}}
    \global\long\def\arraystretch{0.9}%
    \resizebox{.85\textwidth}{!}{{{}}%
    \begin{tabular}{l>{\centering}c>{\centering}c>{\centering}c>{\centering}c>{\centering}c>{\centering}c>{\centering}c>{\centering}c>{\centering}c>{\centering}c>{\centering}c}
    \toprule
		& \multicolumn{8}{c}{Unique network} & \multicolumn{3}{c}{Ensemble of two networks} \tabularnewline
	\cmidrule(lr){2-9}\cmidrule(lr){10-12}
	    & CE & F & SL & JO & ELR & Me & P & DSOS & ELR+ & DSOS & DM  \tabularnewline
    \midrule
       Clothing1M & $69.10$ & $69.84$ & $71.02$ & $72.16$ & $72.87$ & $73.47$ & $73.49$ & \textbf{73.63} & $74.05$ & $74.13$ & \textbf{74.76} \tabularnewline
    \bottomrule
    \end{tabular}}
\end{table*}

\begin{table}[]
\caption{\label{tab:speed} Wall-clock training time comparison for state-of-the-art algorithms on the mini-Webvision dataset. All algorithms were run on an RTX 2080 Ti GPU using the PyTorch \protect\cite{2019_NeurIPS_Pytorch} framework.}
\centering{}
\global\long\def\arraystretch{0.9}%
\resizebox{1\columnwidth}{!}{{{}}%
\begin{tabular}{l>{\centering}c>{\centering}c>{\centering}c>{\centering}c>{\centering}c}
\toprule
 & {M} & {ELR} & {DSOS} &  {ELR+} & {DM}  \tabularnewline
 \midrule
 Epoch & \textit{9.5min} & \textit{10.5min} & \textit{11.25min}  & \textit{28min} & \textit{50min} \tabularnewline
 Full training & \textit{15.75h} & \textit{17.5h} & \textit{18.75h}  & \textit{46.75h} & \textit{83h}   \tabularnewline
 \bottomrule
\end{tabular}}
\end{table}
\subsection{Comparison against the state-of-the-art}
Table~\ref{tab:sotastan} reports results for DSOS when compared to state-of-the-art approaches on the web-corrupted versions of Stanford Cars and MiniImageNet~\cite{2020_ICML_MentorMix}.
Table~\ref{tab:sota} compares DSOS against state-of-the-art algorithms on the WebVision 1.0 dataset~\cite{2017_arXiv_WebVision} reduced to the 50 first classes (mini-WebVision, 66K images), a large scale dataset created using web queries.
Table \ref{tab:sotaclothing} reports results for Clothing1M. When necessary, we differentiate between methods using a unique network for inference and methods using an ensemble of two networks. In this case, we ensemble two networks trained using DSOS from different random initialization and show the direct benefits of using an ensemble in the web label noise scenario.
We compare with loss or label correction algorithms: Forward correction (\textbf{F})~\cite{2017_CVPR_ForwardLoss}, Bootstrapping (\textbf{B})~\cite{2015_ICLR_Bootstrapping}, Probabilistic correction (\textbf{P})~\cite{2019_PCorr_CVPR}, Joint Optimization (\textbf{JO})~\cite{2018_CVPR_JointOpt}, S-Model \textbf{(\textbf{SM})}~\cite{2017_ICLR_Smodel}; sample selection algorithms: Co-Teaching (\textbf{Co-T})~\cite{2018_NeurIPS_CoTeaching}, MentorMix (\textbf{MM})~\cite{2020_ICML_MentorMix}, MentorNet (\textbf{MN})~\cite{2018_ICML_MentorNet}; semi-supervised correction algorithm: DivideMix (\textbf{DM})~\cite{2020_ICLR_DivideMix}, Early Learning Regularization (\textbf{ELR} and \textbf{ELR+})~\cite{2020_NeurIPS_EarlyReg}; regularization algorithms: Mixup (\textbf{M})~\cite{2018_ICLR_mixup}, Symetric cross-entropy Loss (\textbf{SL})~\cite{2019_ICCV_SL}; meta-learning algorithms: Learning to learn (\textbf{Me})~\cite{2019_Meta_CVPR}; standard cross-entropy training (\textbf{CE}), standard cross-entropy plus dropout (\textbf{D}).

\subsection{Training speed}
Table \ref{tab:speed} reports the wall-clock training time for state-of-the-art methods on the mini-Webvision subset. The first line reports average epoch time, warm-up included, and the second line reports the full training duration (100 epochs). Both of these metrics exclude evaluation on a validation set.  We compare against state-of-the-art algorithms performing the best on mini-Webvision~\textbf{DM} \cite{2020_ICLR_DivideMix}, \textbf{ELR} and \textbf{ELR+} \cite{2020_NeurIPS_EarlyReg}, \textbf{M} \cite{2018_ICLR_mixup}. DSOS improves accuracy results on mini-Webvision and trains significantly faster then the closest performing algorithms. Note that the training time for DivideMix \cite{2020_ICLR_DivideMix} heavily depends on the training scenario as the algorithm oversamples the unlabeled data every epoch, i.e.~the epoch length depends on clean/noisy detection.

\subsection{Discussion}
DSOS improves accuracy results on web crawled datasets such as mini-WebVision (Table~\ref{tab:sota}) or web corrupted datasets: miniImageNet (large grained) and Stanford cars (fine grained) in Table~\ref{tab:sotastan}. 
We explain the lower performance on Clothing1M by the specificity of the gathering process for the dataset which, according to the authors~\cite{2015_CVPR_Clothing1M}, contains very high levels of in-distribution noise because the dataset was crawled from a clothes database exclusively. This goes against our hypothesis in Section~\ref{sec:expstudy}. Even then, our results are competitive and convergence is reached faster for DSOS, see Table \ref{tab:speed}.

\section{Conclusion}
\ctB{
This paper provides evidence of the nature of noise (dominantly out-of-distribution) in web-crawled datasets, which we believe to be the reason why improvements reported by recent state-of-the-art noise robust algorithms do not translate to real world noisy datasets. To train a noise-robust neural network on web crawled datasets, we propose DSOS, a simple algorithm using a novel noise detection metric capable of differentiating between clean, in-distribution noisy and out-of-distribution samples. We propose to detect and treat in-distribution and out-of-distribution noise differently to promote a dynamic rejection of unseen out-of-distribution samples, which in turn improves the generalization capabilities of the network.}
DSOS is a much simpler approach to label noise than the top state-of-the-art algorithms that we compare against as we use a one network and online correction strategy with a single training cycle. By properly identifying and correcting the two distinct label noise distributions, DSOS improves on the most competitive state-of-the-art algorithms.
Other strategies could be used to improve network generalization by using out-of-distribution samples such as self-supervised learning, which can learn visual concepts without labels or data augmentation strategies using out-of-distribution samples to augment in-distribution samples. We leave this observation for future work. 

\section*{Acknowledgments} This publication has emanated from research conducted with the financial support of Science Foundation Ireland (SFI) under grant number SFI/15/SIRG/3283 and SFI/12/RC/2289\_P2.

{\small
\bibliographystyle{ieee_fullname}
\bibliography{egbib}
}

\end{document}

\title{Supplementary material for Addressing out-of-distribution label noise in webly-labelled data}

\author{Paul Albert, 
Diego Ortego, Eric Arazo, Noel E. O'Connor, Kevin McGuinness \\
\\
School of Electronic Engineering, \\
Insight SFI Centre for Data Analytics, Dublin City University (DCU) \\
{\tt\small paul.albert@insight-centre.org}
}

\maketitle

\section{Additional explanation of the behavior of the ID/OOD measure}
\begin{figure}
\centering
\includegraphics[width=.95\linewidth]{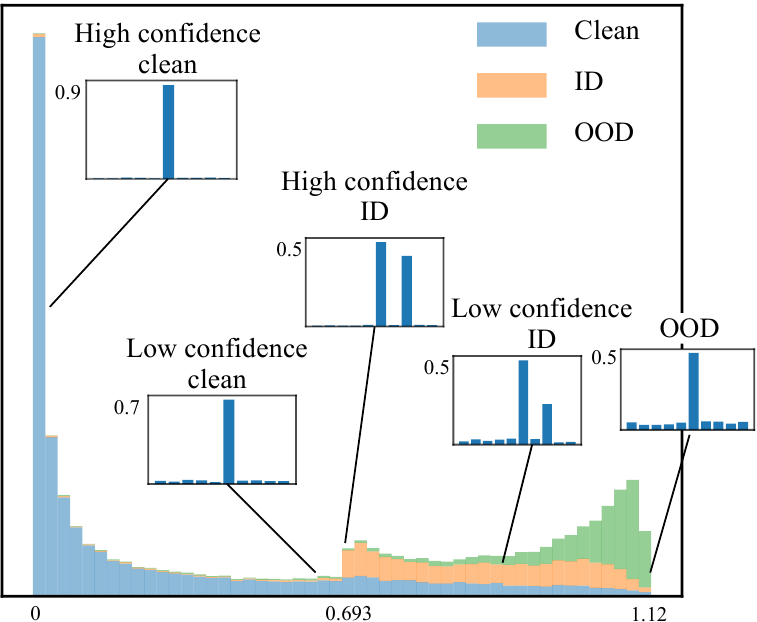}
\par
\caption{Clean/ID/OOD hierarchy established by the collision entropy of the interpolated label $\protect\mathit{l}_\mathit{detect}$\label{fig:interlabmet}}
\end{figure}

Figure~\ref{fig:interlabmet} illustrates the behavior of our proposed metric $\mathit{l}_\mathit{detect}$. By studying the collision entropy of the interpolated label $y_\mathit{inter}$ between the network prediction and the ground truth label, we establish a hierarchy from clean to ID noise to OOD noise. The pivot point $-\log(.5) = 0.693$ marks the separation between low confidence clean samples and high confidence noisy samples. Although some clean samples will be detected as noisy at the pivot point, because we avoid OOD sample during this transition, we can correct the detected confident ID samples without concerns of labeling OOD data or corrupting the clean samples since we relabel correct but simply under-confident clean samples: this will not harm the training procedure (their label stays the same). By smoothing OOD samples, we also avoid correcting ID noisy samples with an under-confident corrected prediction.

\begin{table*}
\caption{\label{tab:hyper} Hyperparameter variations across experiments. We do not change hyperparameters across noise levels for CIFAR-100, miniImageNet, and Stanford Cars.}
\centering{}
\global\long\def\arraystretch{0.9}%
\resizebox{0.9\textwidth}{!}{{{}}%
\begin{tabular}{l>{\centering}c>{\centering}c>{\centering}c>{\centering}c>{\centering}c}
\toprule
 & CIFAR-100 &{Stanford Cars} & {miniImageNet} & {Webvision} & {Clothing1M} \tabularnewline
\midrule
{Network} & PreActResNet18 & InceptionResNetV2 & InceptionResNetV2  & InceptionResNetV2 & ResNet50 \tabularnewline
{ImageNet pretraining} & No & No & No & No & Yes \tabularnewline
{Number of epoch} & $100$ &$400$ & $200$  & $100$ & $100$ \tabularnewline
{Batch size} & $32$ &$32$ & $32$  & $32$ & $32$ \tabularnewline
{Initial learning rate} & $0.03$ & $0.05$ & $0.01$  & $0.01$ & $0.002$ \tabularnewline
{Lr reduction} & $[50,80]$&$[200,300]$ &$[100, 160]$ & $[50, 80]$ & $[50, 80]$ \tabularnewline
{Weight decay} & $5e-4$ & $5e-4$ & $5e-4$ & $5e-4$ & $1e-3$ \tabularnewline 
{Resize} & $32$ & $320$ & $320$ & $256$ & $256$  \tabularnewline 
{RandomResize Range} & $-$ & $[0.75, 1.33]$ & $-$ & $-$ & $-$ \tabularnewline 
{Crop} & $32$ & $299$ & $299$ & $227$ & $224$  \tabularnewline 
{Dropout ratio} & $0.0$ &$0.3$ & $0.0$ & $0.0$ & $0.0$  \tabularnewline 
{$\alpha$}  & $0.05$ & $0.05$ & $0.05$ & $0.05$ & $0.05$ \tabularnewline
{Epoch start correction} & $51$ & $201$ & $101$ & $51$ & $1$ \tabularnewline 
\bottomrule
\end{tabular}}
\end{table*}

\section{Hyperparameter table}
Table \ref{tab:hyper} details the hyperparameters used in every experiment reported in the state-of-the art comparison. The configuration remains the same across different noise ratios for miniImageNet and Stanford Cars. The parameters common to all experiments are: entropy regularization \cite{2018_CVPR_JointOpt}, SGD optimizer, a learning rate decay factor of 10, random horizontal flips, mixup \cite{2018_ICLR_mixup} data augmentation. To match the baseline of \cite{2020_ICML_MentorMix}, we add a dropout layer before the fully connected layer in the case of the Stanford Cars experiments. We do not use dropout for other datasets as we manage to match the baselines without it.

\begin{algorithm*}[t]
\caption{DSOS \label{alg:algo}}
\textbf{Input}: $\mathcal{D} = \left\{ \left(x_{i},y_{i}\right)\right\} _{i=1}^{N} $ a web noise dataset. $h$ at convolutional neural network. \\
\textbf{Parameters}: $\alpha, e_{warmup}, e_{max}$\\
\textbf{Output}: Trained neural network $h_{\phi}$
\begin{algorithmic}[1] %[1] enables line numbers
\For{$e = 1, \dots e_{warmup}$} \Comment{Warmup}
\For{$t=1, \dots numBatches$}
\State Sample the next mini-batch $(x,y)$ from $\mathcal{D}$
\State $L = \mathit{CrossEntro}(h(x_{mixed}), y_{mixed})$
\State $\mathit{UpdateNetworkWeights}($L$)$
\EndFor
\EndFor
\For{$e = e_{warmup}+1, \dots e_{max}$} \Comment{Label correction}
\State $\tilde{U}, \tilde{V}, \mathit{predictions} = \mathit{EvaluateMetrics}(h, \mathcal{D})$ \Comment{Evaluated with regards to the original labels}
\For{$y_i = y_1, \dots y_N$}
\If{$\tilde{u}_i > 0.9$}  \Comment{In-distribution bootstrapping, $\tilde{U} =  \{\tilde{u}_{i}\} _{i=1}^{N}$}
\State $y_i = p_i$ \Comment{$\mathit{predicitions} =  \{p_{i}\} _{i=1}^{N}$}
\EndIf
\State $y_i = \mathit{Softmax}(y_iv_i/\alpha)$ \Comment{Dynamic Softening, $\tilde{V} =  \{\tilde{v}_{i}\} _{i=1}^{N}$}
\EndFor
\For{$t=1, \dots \mathit{numBatches}$}
\State Sample the next mini-batch $(x,y)$ from $\mathcal{D}$ \Comment{Train on the corrected labels}
\State $\tilde{V}_{mini}$ the values in $\tilde{V}$ for the samples in the mini-batch
\State $L = \mathit{CrossEntro}(h(x), y)$
\State $L = L + 0.4\times\mathit{EntroPen}(h(x), \tilde{V}_{mini})$ \Comment{Weighted entropy penalization}
\State $\mathit{UpdateNetworkWeights}($L$)$
\EndFor
\EndFor
\State \textbf{return} $h$ \Comment{Robustly trained network}
\end{algorithmic}
\end{algorithm*}

\begin{figure*}
\centering
\includegraphics[width=.85\linewidth]{"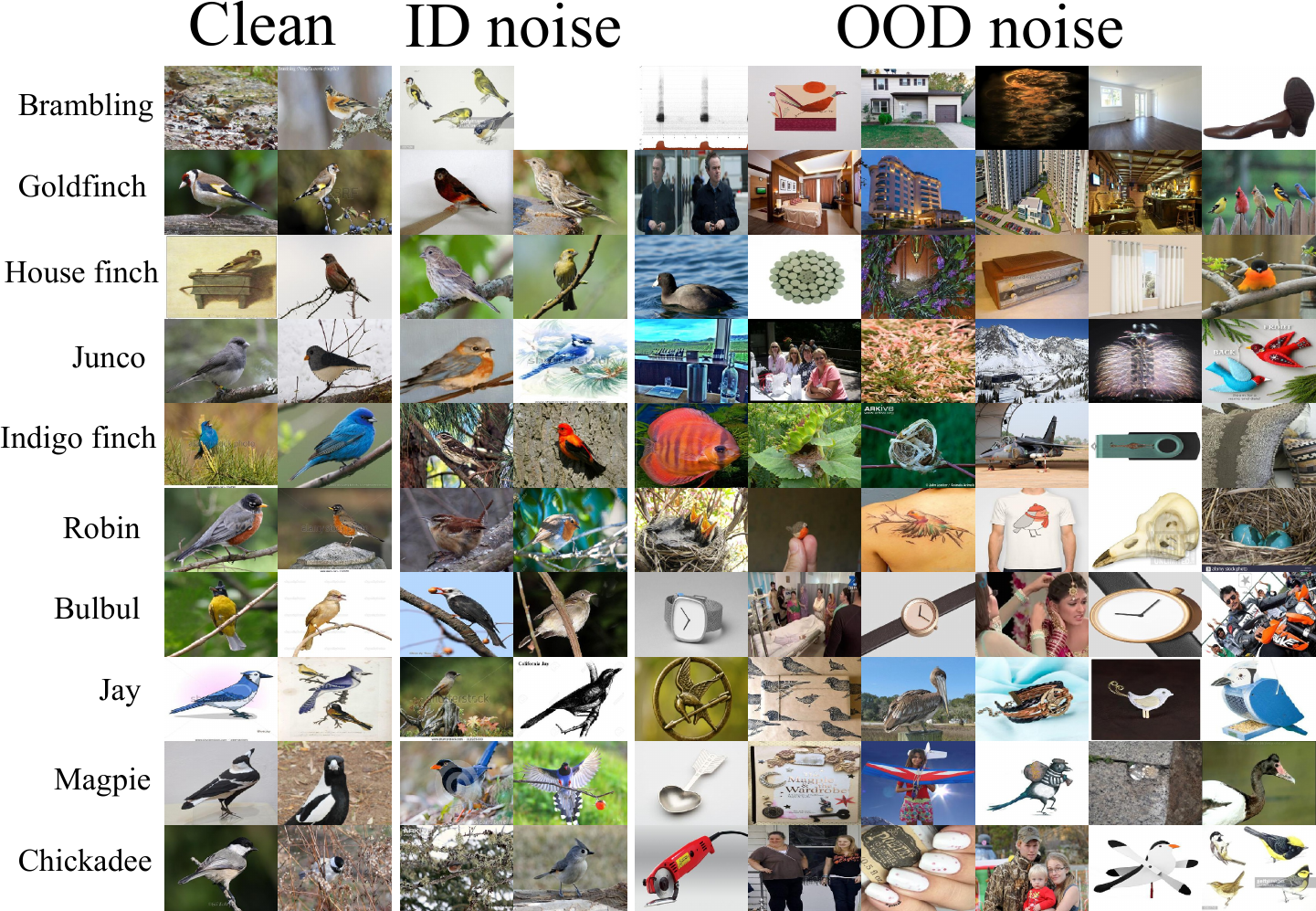"}
\par
\caption{Samples annotated as clean, in-distribution noise, out-of-distribution noise.\label{fig:Noiseimages1}}
\end{figure*}

\begin{figure*}
\centering
\includegraphics[width=.85\linewidth]{"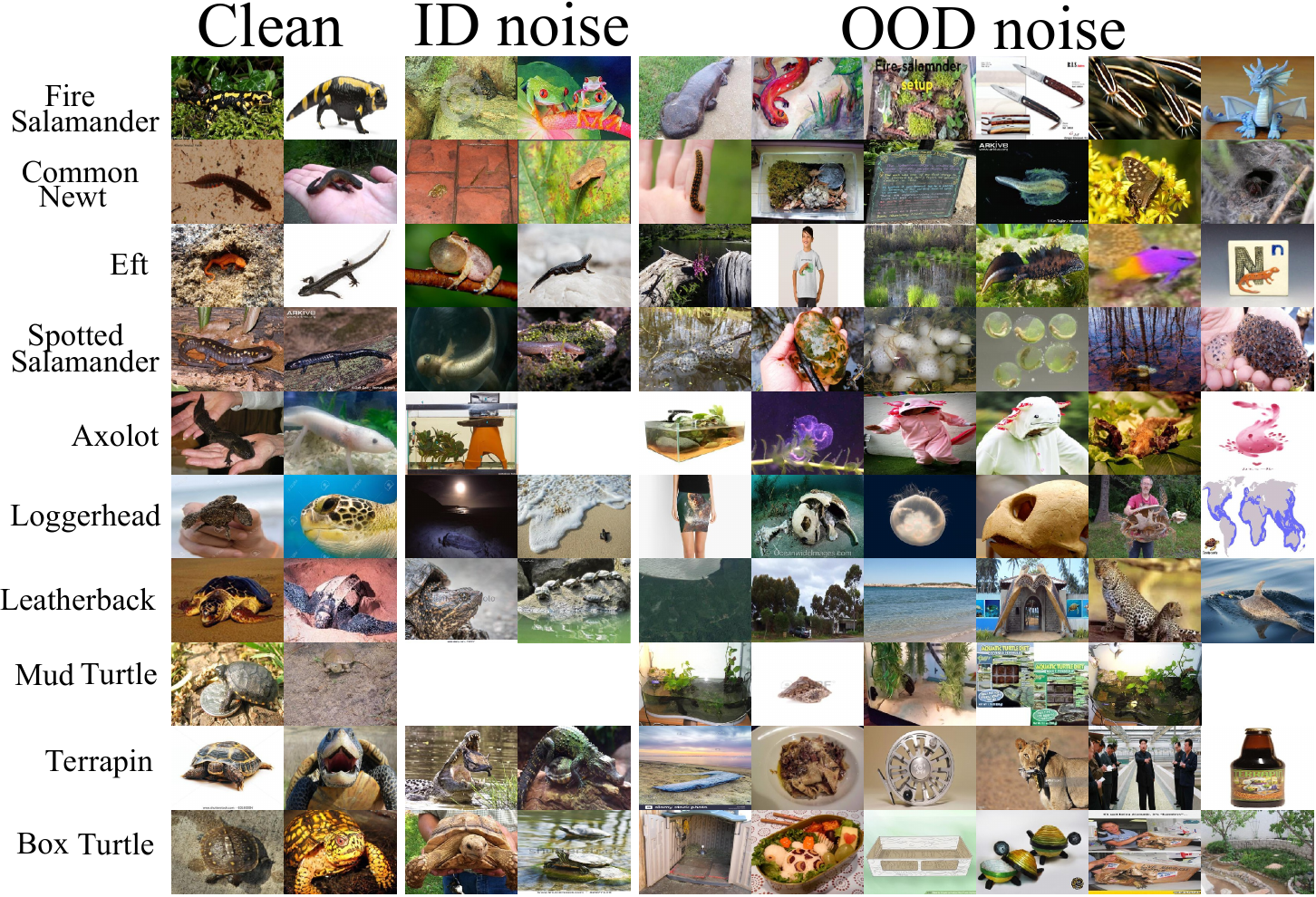"}
\par
\caption{Samples annotated as clean, in-distribution noise, out-of-distribution noise.\label{fig:Noiseimages2}}
\end{figure*}

\section{Algorithm}
Alg. \ref{alg:algo} displays the DSOS algorithm.

\section{Examples of labeled images from the mini-Webvision subset}
Figures \ref{fig:Noiseimages1} and \ref{fig:Noiseimages2} display examples of images labeled from the mini-Webvision subset. The annotations are available together with our code at [github].

{\small
\bibliographystyle{ieee_fullname}
\bibliography{egbib}
}